\pdfoutput=1

\documentclass[11pt]{article}

\usepackage[final]{coling}

\usepackage{times}
\usepackage{latexsym}

\usepackage[T1]{fontenc}

\usepackage[utf8]{inputenc}

\usepackage{microtype}

\usepackage{inconsolata}

\usepackage{inconsolata}
\usepackage{array,multirow,graphicx}
\usepackage{wrapfig}

\newcommand{\revised}[1]{{\color{black} #1}}

\title{\vspace{-1.25em}\hbox{\includegraphics[height=0.85em, trim={-12em 24.5em 0 -14em}]{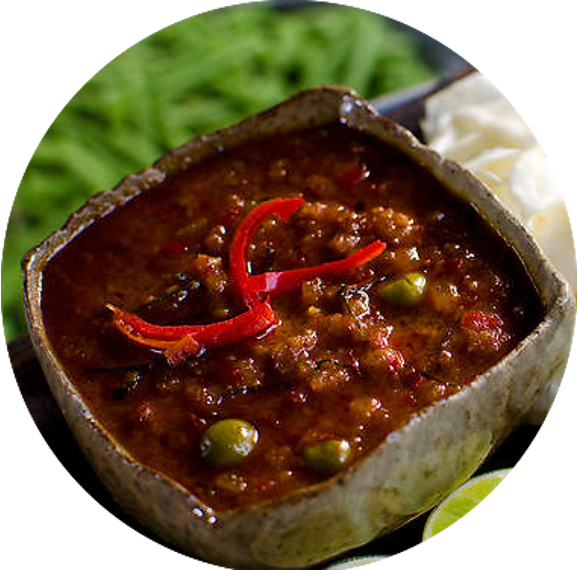}} PrahokBART: A Pre-trained Sequence-to-Sequence Model \\for Khmer Natural Language Generation}

\author{
\textbf{Hour Kaing}, \ \ 
\textbf{Raj Dabre}, \ \ 
\textbf{Haiyue Song}, \ \ 
\\
\textbf{Van-Hien Tran}, \ \ 
\textbf{Hideki Tanaka}, \ \ 
\textbf{Masao Utiyama}, \ \ 
\\
National Institute of Information and Communications Technology, Kyoto, Japan\\
{\tt \{hour\_kaing, raj.dabre, haiyue.song\}@nict.go.jp} \\ 
{\tt \{tran.vanhien, hideki.tanaka, mutiyama\}@nict.go.jp}
\\
\hbox{\includegraphics[height=0.75em, trim={0em 0.5em -1em -0.2em}]{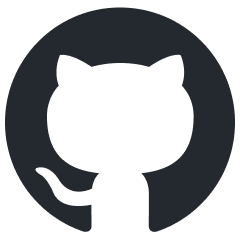}}: \url{https://github.com/hour/prahokbart}
\\
\hbox{\includegraphics[height=0.75em, trim={0em 5em -2em 2em}]{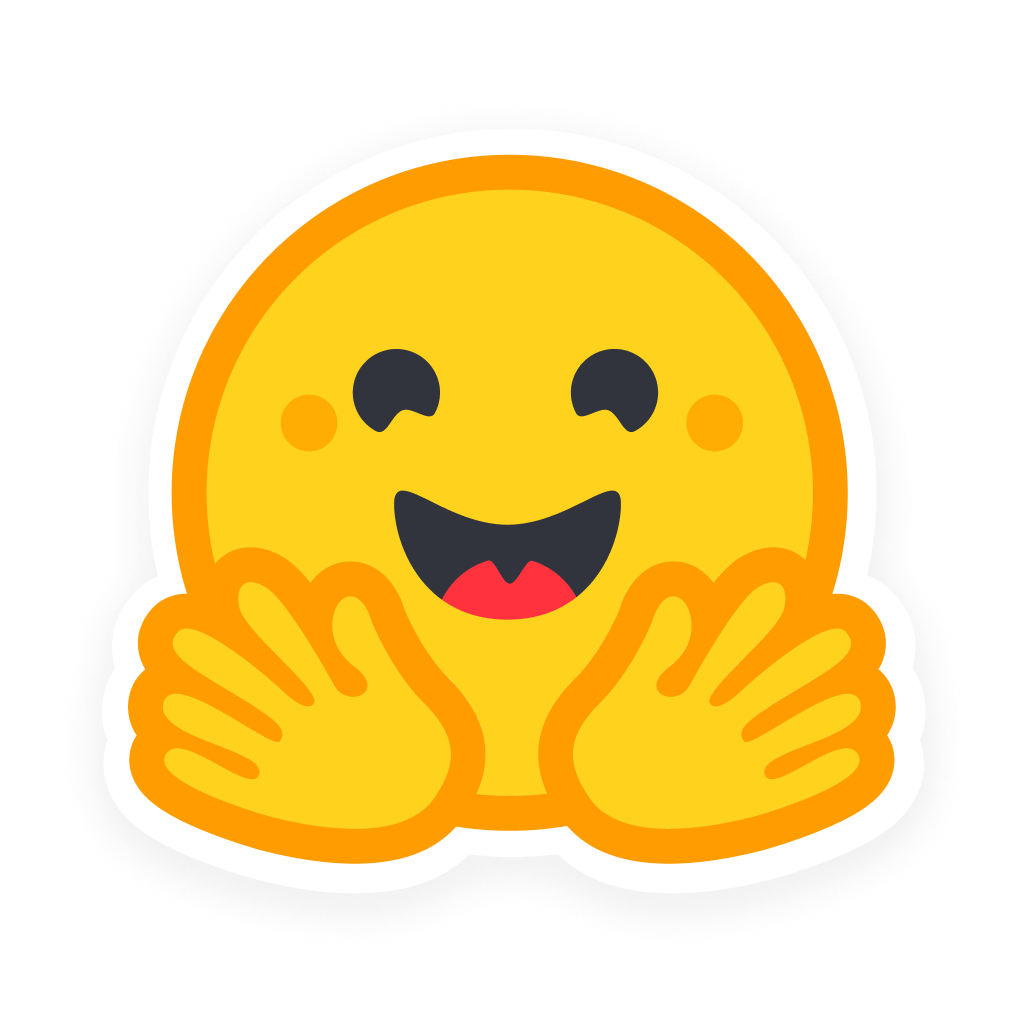}}: \url{https://huggingface.co/prajdabre/prahokbart}
}

\begin{document}
\maketitle
\begin{abstract}
This work introduces {\it PrahokBART}, a compact pre-trained sequence-to-sequence model trained from scratch for Khmer using carefully curated Khmer and English corpora. 
We focus on improving the pre-training corpus quality and addressing the linguistic issues of Khmer, which are ignored in existing multilingual models, by incorporating linguistic components such as word segmentation and normalization.
We evaluate PrahokBART on three generative tasks: machine translation, text summarization, and headline generation, where our results demonstrate that it outperforms mBART50, a strong multilingual pre-trained model. 
Additionally, our analysis provides insights into the impact of each linguistic module and evaluates how effectively our model handles space during text generation, which is crucial for the naturalness of texts in Khmer.
\end{abstract}

\section{Introduction}

Pre-trained sequence-to-sequence (PS2S) models have been proven to be data-efficient and effective in enhancing performance across various natural language generation (NLG) tasks, including machine translation, text summarization \cite{lewis2020bart}, and headline generation \cite{sarti2024it5}. These models are typically pre-trained on extensive raw text corpora using denoising objectives and fine-tuned on task-specific data, as seen with models like BART \cite{lewis2020bart}. Recently, many PS2S models have been developed as multilingual, with models like mBART50 \cite{tang2020multilingual} and mT5 \cite{xue2021mt5} being trained on over fifty languages simultaneously. 
Such multilingual PS2S models have been particularly advantageous for low-resource languages (LRLs), as they can leverage linguistic similarities with high-resource languages (HRLs) \cite{10.1145/3406095}. Improvements in LRLs are often observed when they share linguistic features with HRLs, such as similar syntax \cite{ahmad2019difficulties}, overlapping vocabularies \cite{patil2022overlap}, and code-switching \cite{pires2019multilingual}.

\begin{figure}
    \centering
    \includegraphics[trim={0 0 0 0},clip,width=1\linewidth]{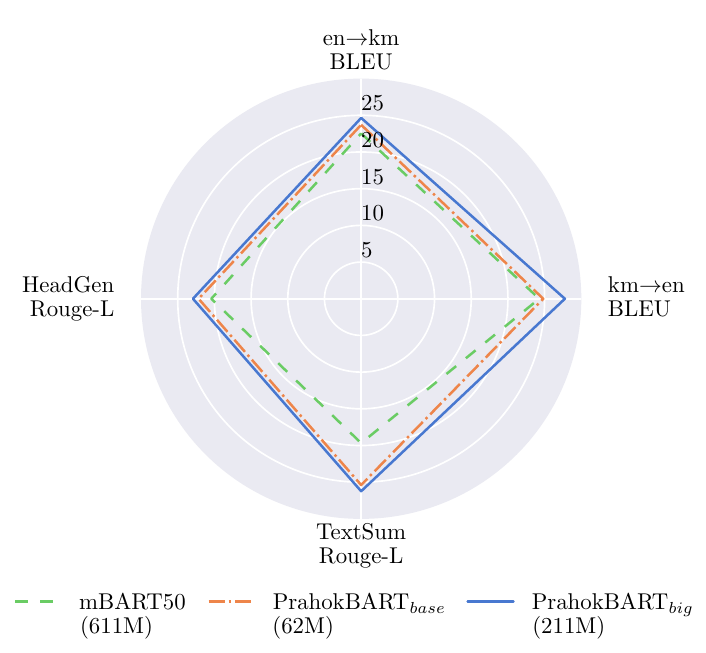}
    \caption{Overall performance of PrahokBART across tasks.}
    \label{fig:plot_results}
    \vspace{-3mm}
\end{figure}

Despite the advantages of multilingual PS2S models, they face significant challenges due to the need for vast model parameters to accommodate a large and linguistically diverse corpus. This often leads to the under-representation of languages with scarce resources or unique linguistic features, such as distinctive writing systems. To address these issues, researchers have developed language-specific \cite{eddine2021barthez, nguyen2022bartpho, araujo2024sequence} and language group-specific \cite{dabre2022indicbart, reid2021afromt} PS2S models, which are smaller and offer higher performance in downstream tasks compared to their multilingual counterparts. However, many low-resource languages, particularly those with unique writing systems and minimal vocabulary overlap with other languages, like Khmer, still lack these specialized~models.

Research on NLG for the Khmer language is scarce and predominantly limited to multilingual studies \cite{tang2020multilingual, xue2021mt5, costa2022no, palen2022lr}. While these studies have made significant strides, they often overlook the linguistic challenges posed by Khmer, such as the absence of word boundaries \cite{buoy2021joint, kaing2021towards}, encoding ambiguities \cite{hosken2022khmer}, and linguistic roles of spaces. Although there is no standard way or requirement of spacing in Khmer texts, native speakers commonly use them between phrases or sentences, as a comma, or just for readability, which make texts more natural. Let's denote such spaces as {\it functional spaces}.
All these issues are frequently ignored due to the reliance on language-agnostic techniques such as the SentencePiece subword tokenizer which treats all characters indiscriminately, which are not tailored to the specific needs of Khmer. This raises critical questions: 1) \textit{do linguistic modules like word segmentation and normalization still hold value in PS2S for Khmer?} 2) \textit{how well do the current models generate functional spaces?}

This work addresses the aforementioned issues by introducing a language-specific PS2S model for the Khmer language, {\it PrahokBART},\footnote{``Prahok is a salted and fermented fish paste used in Cambodian cuisine as a seasoning or a condiment.''---Wikipedia. \url{https://en.wikipedia.org/wiki/Prahok}}\footnote{\revised{The figure near our paper's title is cropped and originally from \url{https://commons.wikimedia.org/wiki/File:Prahokktis.jpg}}} in combination with linguistic modules such as normalization and word segmentation. 
Normalization ensures consistency and uniformity in the texts, making the corpus more predictable for the model to learn from. 
Word segmentation, applied before subword tokenization, ensures that the resulting subword tokens are more linguistically motivated and meaningful units. Additionally, our word segmentation module preserves functional spaces, treating them as individual tokens to enhance model learnability \cite{gow2022improving}.
Essentially, our model trained on carefully curated Khmer and English corpora is more compact and computationally efficient compared to its multilingual counterpart, mBART50. 
We evaluate PrahokBART on three generative tasks: machine translation, text summarization, and headline generation. Our experiments demonstrate that PrahokBART outperforms other models across all tasks as in Figure \ref{fig:plot_results}.


Our analysis highlights the essential role each module plays in enhancing the performance of our model. Additionally, a thorough evaluation of the functional spaces generated by our model, compared to baseline systems, demonstrates that our approach produces these functional spaces with superior quality and accuracy. By addressing these linguistic nuances, our model becomes a more effective PS2S model, adeptly managing the complexities of the Khmer language.

Our key contributions are as follows:
\begin{itemize}
    \item We introduce PrahokBART, the first compact PS2S model specifically designed for the Khmer language, incorporating essential linguistic modules such as normalization and word segmentation.
    \item We evaluate our model on three generative tasks---machine translation, text summarization, and headline generation---and demonstrate that it outperforms the multilingual mBART50 model in terms of efficiency and generation quality measured by BLEU, ChrF, and Rouge-L.
    \item We analyze the impact of each linguistic module and assess the quality of the functional spaces generated by our model. Our findings indicate that word segmentation and normalization are crucial for PS2S models, particularly for languages with characteristics similar to Khmer.
\end{itemize}


\section{Related Works}

{\bf Pre-trained Models}: Pre-trained models have revolutionized the field of natural language processing (NLP). \citet{devlin2019bert} introduced BERT, an encoder-only model designed for natural language understanding (NLU). 
\revised{Although decoder-only models, such as GPT \cite{adelani2022few} and other recent large models that support Khmer \cite{touvron2023llama, nguyen2023seallms}, perform well across various natural language generation (NLG) tasks}, encoder-decoder models, despite being more compact, have been proven to be the most effective for NLG \cite{radford2019language, tay2023ul}. Notable state-of-the-art models in this category include BART \cite{lewis2020bart} and T5 \cite{raffel2020exploring}. These models have been further extended to multilingual settings, with examples such as mBERT \cite{devlin2019bert}, XLM-R \cite{conneau2020unsupervised}, mBART50 \cite{tang2020multilingual} and mT5 \cite{xue2021mt5}, benefiting many languages simultaneously including Khmer.


{\bf Language-Specific Pre-trained Models}: Research has also focused on pre-trained models tailored to specific languages or language groups. For natural language understanding (NLU), there are models designed for French \cite{martin2020camembert}, Vietnamese \cite{nguyen2020phobert}, Indian \cite{kakwani2020indicnlpsuite}, and others. For natural language generation (NLG), many models exist for French \cite{eddine2021barthez}, Vietnamese \cite{nguyen2022bartpho}, Spanish \cite{araujo2024sequence}, Indonesian \cite{cahyawijaya2021indonlg}, African \cite{ogundepo2022afriteva, adelani2022few, meyer2024nglueni}, Indian \cite{dabre2022indicbart,j-etal-2024-romansetu}, and many more. 
\revised{Notably, models for Indonesian, African, and Indian languages are multilingual but are designed for specific language groups.}
\citet{jiang2021pretrained} pre-trained a BERT model for Khmer, and evaluated the model on NLU tasks such as POS tagging and document classification. 
To the best of our knowledge, there is no work on NLG pre-trained models specified for Khmer.

{\bf Language-Specific NLG Benchmarks}: There are NLG benchmarks for specific languages such as Indian \cite{kumar2022indicnlg, dixit2023indicmt}, Indonesian \cite{cahyawijaya2021indonlg}, French \cite{eddine2021barthez}, Vietnamese \cite{nguyen2022bartpho}, and many African languages \cite{adebara2024cheetah, adelani2022few, meyer2024nglueni}. However, there is no formal NLG benchmark for Khmer, and the datasets accumulated in this paper could be used as a benchmark.

\section{PrahokBART}
We now describe the design behind PrahokBART.
\subsection{Data Curation}

{\bf Data Sources}: We collect pre-training data in Khmer and English from two public sources: Common Crawl (CC) and Wikimedia (Wiki). We include English data in our pre-training process because Khmer texts often contain English words \revised{in their original Latin script form}, primarily proper nouns. For Khmer data, we utilize CC data from mC4 \cite{raffel2020exploring} and WMT2020 \cite{loic2020findings}, and extracte high-quality Khmer content from Wikipedia and Wikibooks.\footnote{${\tt 20240101}$ version of Wikimedia dumps.} For English, we use CC data from mC4. Given that English data is much more abundant than Khmer data, we sample a portion that is five times larger than the Khmer data to balance the datasets. 
Additionally, the combined data includes both document-level and sentence-level content, particularly for Khmer, because the WMT2020 monolingual corpus is sentence-based. 
The reason we retain the English data at a size five times larger than the Khmer data is due to the scarcity of high-quality Khmer data for pre-training. Including more English data helps to better generalize the model during pre-training. Similarly, the mix of document-level and sentence-level data, particularly for Khmer, is maintained for the same reason: to maximize the available data and enhance the robustness of the model.

{\bf Data Cleaning}: This step aims to minimize noisy texts that could negatively impact the learning capability of pre-trained models. It involves normalization, filtering, and removal of excessive spaces. We apply normalization to all texts to prevent the loss of high-quality content; details of this process are provided in Section \ref{sec:preproc}. We also find that some \revised{Khmer} texts, particularly from Common Crawl (CC), were tokenized with spaces as word delimiters. While we cannot trace the exact source, these texts likely originated from preprocessed corpora. 
Additionally, the functional spaces in those texts were eliminated perhaps by a particular segmenter. 
We do not need those word-delimiter spaces and remove them, resulting in zero spacing in those texts. We identify texts that contain word-delimiter spaces using a ratio of space-to-character whether their ratios are larger than $0.2$.\footnote{We manually checked several samples and found those with a space ratio of $0.2$ to be more natural.}
Furthermore, \revised{both Khmer and English} texts are filtered according to the rules listed in Table \ref{tab:filter_constraints}, following the approach of \citet{costa2022no}\footnote{\revised{The rule of ``Ratio of functional spaces > $32\%$'' has no effect on Khmer texts due to our removal of delimiter spaces.}}. The cleaned dataset for pre-training consists of approximately \revised{$4.2$ billion tokens: $0.7$ billion tokens for Khmer and $3.5$ billion tokens for English.}

\begin{table}[t!]
    \centering
    \begin{tabular}{lr}
        \revised{Filtering Rules} & \\
        \hline
        Number of characters & $<10$ \\
        Number of time characters repeated & $>20$ \\
        Ratio of functional spaces & $>30\%$ \\
        Ratio of numbers & $>20\%$ \\
        Ratio of emojis & $>10\%$ \\
        Ratio of punctuation & $>20\%$ \\
        \hline
        Ratio of unmatched scripts & $>5\%$ \\
        Probability of being target language & $<50\%$ \\
        \hline
    \end{tabular}
    \caption{\revised{Rules for filtering noisy documents: A document is removed if any of these rules are satisfied.} We used the language identifier used for NLLB \cite{costa2022no} to compute the probability of being target language.}
    \label{tab:filter_constraints}
    \vspace{-3mm}
\end{table}

\subsection{Preprocessing}\label{sec:preproc}

\begin{figure}
    \centering
    \includegraphics[trim={20 10 0 0},clip,width=0.97\linewidth]{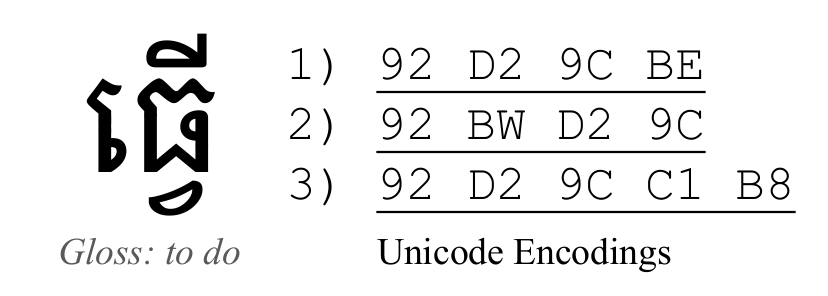}
    \caption{Example of ambiguous encodings. Three sequences on the right with different last two hexadecimal Unicode values represent the same word on the left. These Unicode values all start with {\tt U+17}.}
    \label{fig:encoding_exmaple}
    \vspace{-3mm}
\end{figure}

{\bf Normalization}: This step consists of invisible characters removal ({\tt rm\_inv}) and encodings normalization ({\tt enc\_norm}). Khmer texts use complex scripts that can lead to encoding ambiguities \cite{hosken2022khmer}, which can adversely affect NLP models, including machine translation systems \cite{kaing2024robust}. 
An example in Figure \ref{fig:encoding_exmaple} is a word that can be represented by different encodings. The first sequence aligns with the word's spelling and is typically used by typists. The other two sequences might be chosen occasionally based on the typist's convenience. For example, a typist might use {\tt C1} and {\tt B8} instead of {\tt BE} if they are more familiar with the keyboard positions of {\tt C1} and {\tt B8}.
Beside the ambiguous encodings, we find that invisible characters, such as zero-width white-spaces, are frequently used in Khmer corpora. These characters often control text appearance \cite{hosken2022khmer} or serve as word separators to improve text display, particularly on web pages. We believe that these invisible characters are generally unnecessary for NLP models, with the exception of cases where text visuals depend on the zero-width white-spaces. However, such cases are rare and can be addressed using dictionaries or specific rules. Consequently, we remove invisible characters and apply normalization rules as outlined by \citet{hosken2022khmer}. For a best practice, {\tt rm\_inv} need to be applied before {\tt enc\_norm} because normalization rules do not consider invisible characters and could be broken by the presence of the invisible characters. \revised{Implementation details are explained in Appendix \ref{implement_norm}}.

{\bf Word Segmentation}: This step segments a text into a sequence of words, typically applied before subword tokenization or during the pretokenization stage \cite{mielke2021between}, particularly for languages without explicit word boundaries. Word segmentation is optional when using a Unigram tokenizer \cite{kudo2018subword}. Consequently, many NLP systems, especially multilingual ones, often omit the word segmentation module to simplify the pipeline. However, without word segmentation, a frequency-based Unigram tokenizer might merge separate words or parts of words into a single subword, which can be semantically meaningless.
As an example, consider a string like `regret at his' without delimiter spaces between words as `regretathis'. A Unigram tokenizer might transform this into a sequence of subwords such as `regret a this'. Apparently, the tokenizer incorrectly combines `t' and `his' into a single subword due to their frequent co-occurrence in the corpus, which make the string meaningless. Therefore, segmenting the string into words beforehand prevents `t' and `his' from being combined. This way will generate more meaningful subword tokens, thereby enhancing the performance of downstream NLP systems \cite{he2020dynamic, song2022bertseg}. 

The word segmenter we used also preserves functional spaces in the text, treating them as individual tokens. This is crucial for a model when generating Khmer texts, as functional spaces are integral to making the text appear natural and readable. Without these spaces, Khmer text can become difficult to read and may seem unnatural. 
Similar to the motivation behind performing word segmentation, treating functional spaces as individual tokens prevents the subword tokenizer from combining them with subwords.
To perform word segmentation, we utilize the {\tt khmer-nltk} toolkit.\footnote{\url{https://github.com/VietHoang1512/khmer-nltk}}

{\bf Subword Tokenization}: In this step, we utilize the Unigram subword tokenizer \cite{kudo2018subword}, which has been widely used, especially for languages without explicit word boundaries. We train a subword tokenizer with a $32$k vocabulary using SentencePiece \cite{kudo2018sentencepiece}, and apply this tokenizer to all models in the experiment, with the exception of the mBART50 model.

\subsection{Model and Training Details}
We introduce two versions of PrahokBART, that is, {\bf base} and {\bf big} models, trained using {\tt YANMTT} toolkit\footnote{\url{https://github.com/prajdabre/yanmtt}} \cite{dabre2023yanmtt}. The {\bf base} model has $6$ encoder and decoder layers with $8$ attention heads, and $512$ and $2048$ dimension of hidden and intermediate layers. We double the attention heads ($16$), hidden ($1024$), and dimensions of intermediate ($4096$) layers for the {\bf big} model. Both models have maximum sequence length of $1024$, which can handle medium long documents. We split documents that exceed this maximum length and splitters are at the beginning of sentences that exceed the maximum length. We use sentence splitter provided by {\tt khmer-nltk}.
For other configuration, we mainly follow that of \citet{dabre2022indicbart}, which masks $35\%$ of the words in each sentence by randomly sampling a span length according to a Poisson distribution ($\lambda = 3.5$), which uses dropouts of $0.1$, label smoothing of $0.1$, Adam optimizer with a maximum learning rate of $0.001$, weight decay of $0.00001$, linear learning rate warm-up and decay with $16,000$ warm-up steps, and batch sizes of $4,096$ tokens. We pre-train both models with approximately $16$ epochs on $40$ NVIDIA~V-$100$~GPUs.

\section{Experiments}

\begin{table*}[t!]
    \centering
    \resizebox{\textwidth}{!}{%
    \begin{tabular}{l|rrr|rrrrrr}
        \hline
         Model & \multicolumn{2}{c}{\#params} & FLOPs & \multicolumn{2}{c}{en$\rightarrow$km} & \multicolumn{2}{c}{km$\rightarrow$en} & TextSum & HeadGen \\
         (initialization)& \multicolumn{2}{c}{(million)} & (billion) & BLEU & ChrF & BLEU & ChrF & Rouge-L & Rouge-L \\
        \hline
        Random & $62$ & $1${$\times$}  & $1.13$ & $19.47$ & $49.07$ & $19.47$ & $45.48$ & $10.67$ & $11.10$ \\
        mBART50 & $611$ & $10${$\times$}  & $9.06$ & $22.53$ & $52.47$ & $24.27$ & $50.32$ & $19.67$ & $20.42$ \\
        \hline
        PrahokBART$_{base}$ & $62$ & $1${$\times$}  & $1.13$ & $23.70$ & $52.51$ & $24.81$ & $49.99$ & $25.38$ & $22.15$ \\
        PrahokBART$_{big}$ & $211$ & $3${$\times$}  & $4.53$ & $\textbf{24.64}$\rlap{$^\dag$} & $\textbf{53.54}$\rlap{$^\dag$} & $\textbf{27.76}$\rlap{$^\dag$} & $\textbf{53.26}$\rlap{$^\dag$} & $\textbf{26.23}$ & $\textbf{22.92}$ \\
        \hline
    \end{tabular}
    }
    \caption{Main results for 3 tasks. \#params indicates number of model parameters including embeddings, FLOPs are per token FLOPs computed following {\it Chinchilla} \cite{hoffmann2022training} without embedding and with the sequence length of one. $^{\dag}$ denotes statistical significance with $p<0.01$ compared with the second best result.}
    \label{tab:main}
    \vspace{-3mm}
\end{table*}

\subsection{Tasks, Datasets and Evaluation}

\textbf{Machine Translation (MT)}: We evaluate our model on the English$\leftrightarrow$Khmer translation task. We use Asian Language Treebank (ALT) dataset \cite{riza2016introduction} following the standard splits.\footnote{\url{https://www2.nict.go.jp/astrec-att/member/mutiyama/ALT/}}
For the English to Khmer direction, we preprocess both the translation output and the references by normalization, word segmentation, and removal of all functional spaces. We remove the functional space to focus only on the translation of the contents. 
For evaluation metrices, we compute BLEU\footnote{BLEU+nrefs:1+case:mixed+eff:no+tok:13a+smooth:exp +version:2.3.1} \cite{papineni2002bleu} and ChrF\footnote{ChrF2+nrefs:1+case:mixed+eff:yes+nc:6+nw:0+space:no +version:2.3.1} \cite{popovic2015chrf} using SacreBLEU \cite{post2018call}. 
We also evaluate this task using COMET\footnote{\url{https://github.com/Unbabel/COMET}} \cite{rei2020comet} scores. In contrast to the evaluation using BLEU and ChrF, we do not preprocess neither translation outputs nor references, that is, word segmentation and functional space removal. COMET relies on XLM-RoBERTa encoder and our preprocessing step will cause its input texts incompatible with its tokenizer. For those translation outputs especially generated by our PrahokBART, we detokenize the subwords and then the words.
\\
\textbf{Text Summarization (TextSum)}: This task is to compress an article into a compact paragraph or summary. 
We use a multilingual dataset, Lr-sum \cite{palen2022lr}, which contains a Khmer dataset, for our experiment. The Lr-sum dataset contain titles, summaries, and body texts. We pair body texts and summaries as input and output for this task.
Similar to the above MT task, we preprocess both output summaries and the references. We use Rouge-L \cite{lin2004rouge}, a wisely used evaluation metric for text summarization, and compute it using a modified version toolkit for multilangual summarization \cite{hasan2021xl}.
\\
\textbf{Headline Generation (HeadGen)}: This task aims to generate a headline or title for an article. In our experiment, the models for this task take a summary as input and generate a title as output.
We also evaluate our model on the Lr-sum dataset \cite{palen2022lr} by pairing its summaries and titles as input and ouput. For evaluation, preprocessing and the evaluation metrics are the same as that of in TextSum.
\revised{Additionally, we conducted a statistical significance test for all tasks using paired bootstrap resampling \cite{koehn2004statistical} with $1$k bootstrap resamples.}

\subsection{Model Fine-tuning and Baselines}

{\bf Fine-tuning}: We fine-tune all the tasks with Adam optimizer, learning rate of $0.001$, dropout rate of $0.1$, label smoothing of $0.1$, warm up steps of $16k$, and weight decay of $10^{-5}$. We fine-tune the model until convergence and validate it every $1$k steps using development set with number of patience of $20$ consecutive validations. We also set maximum length of source-target during fine-tuning to $256$-$256$ for MT, $512$-$64$ for TextSum, and $64$-$32$ for HeadGen.

\noindent{\bf Random}: We train the downstream models with random parameter initialization. This baseline configuration, similar to PrahokBART$_{base}$, has $6$ encoders and $6$ decoders, each with 8 attention heads, and hidden and intermediate layer dimensions of $512$ and $2048$, respectively.

\noindent{\bf mBART50}: This model was pre-trained on 50 languages including Khmer and features deeper encoder and decoder layers (12 layers each) compared to PrahokBART$_{big}$. Other hyperparameters, such as the number of attention heads and the dimensions of hidden and intermediate layers, match those of PrahokBART$_{big}$. Additionally, the vocabulary size of mBART50 is $250$k to accommodate 50 languages, which is seven times larger than that of PrahokBART$_{big}$.

\subsection{Main Results}

\begin{table}[t!]
    \centering
    \begin{tabular}{l|rr}
         Model & en$\rightarrow$km & km$\rightarrow$en \\
        \hline
        Random & $70.51$ & $72.41$ \\
        mBART50 & $74.71$ & $78.47$ \\
        \hline
        PrahokBART$_{base}$ & $76.28$ & $79.36$ \\
        PrahokBART$_{big}$ & $\textbf{77.69}$\rlap{$^\dag$} & $\textbf{82.00}$\rlap{$^\dag$} \\
        \hline
    \end{tabular}
    \caption{Translation results based on COMET. $^{\dag}$ denotes statistical significance with $p<0.01$ compared with the second best result.}
    \label{tab:comet_result}
\end{table}

Table \ref{tab:main} compares the performance of our pre-trained models with mBART50 and Random baselines. 
It is not surprising that all pre-trained models bring a significant improvement compared with the Random baseline even by PrahokBART$_{base}$ with the same number of parameters. We can further see that PrahokBART$_{base}$ has comparable performance with mBART50 even the number of parameters is ten times smaller and lower computational cost in terms of FLOPs.\footnote{\url{https://github.com/karpathy/nanoGPT/blob/master/scaling_laws.ipynb}} Furthermore, by increasing the model size, we boosted the performance across all tasks with PrahokBART$_{big}$, of which the number of parameters is still three times smaller than that of mBART50. However, having a larger model than PrahokBART$_{big}$ might not yield improvement and the number of parameter of PrahokBART$_{big}$ is approximately optimal for the current number of pre-training tokens \cite{hoffmann2022training}. 
The results suggest that there is room for improvement with more parameter-efficient methods compared to fine-tuning on multilingual pre-trained models, such as mBART50, especially for underrepresented languages like Khmer. We further reported COMET scores for the MT task as in Table \ref{tab:comet_result} and showed the superior performance of our models compared with all baselines in terms of the COMET scores. 

\begin{table}[t!]
    \centering\resizebox{\columnwidth}{!}{%
    \begin{tabular}{l|rrr}
         Method & Rouge-1 & Rouge-2 & Rouge-L \\
        \hline
        Lead-3 & $8.03$ & $4.52$ & $7.85$ \\
        LexRank & $7.59$ & $4.57$ & $7.38$ \\
        PrahokBART & $\textbf{30.60}$ & $\textbf{18.00}$ & $\textbf{26.23}$ \\
        \hline
        Oracle & $65.36$ & $57.56$ & $63.51$ \\
        \hline
    \end{tabular}
    }
    \caption{Comparing with two extractive approaches and the oracle for TextSum.}
    \label{tab:compare_extractive}
    \vspace{-3mm}
\end{table}

For TextSum, \citet{palen2022lr} showed that simply taking the first three sentences (Lead-3) produces competitive results for Khmer in their experiment. They also showed that the extractive approach named LexRank \cite{erkan2004lexrank} achieved the best performance for Khmer. Table \ref{tab:compare_extractive} further compares our model with Lead-3 and LexRank. Similar to \citet{palen2022lr}, we included the upper bound scores by selecting a single sentence from that article that has the highest Rouge-L score (Oracle). As a result, our model outperforms both Lead-3 and LexRank. However, TextSum is challenging as the best model is still far behind the oracle. We believe a more advanced technique would introduce a better result. 

\section{Discussion and Analysis}\label{sec:analysis}

Despite the straightforward evaluation of our models on the downstream task above, here we take a deeper look at the success of our models by analyzing the impact of normalization and word segmentation, functional space generation capability of our model, and the impact of pre-training data.

\subsection{Impact of Normalization}

\begin{table}[t!]
    \centering
    \resizebox{\columnwidth}{!}{
    \begin{tabular}{l|rrrr}
         Preprocessing & CC & Wiki & ALT & Lr-sum \\
         \hline
        Original corpus & $4.31$ & $5.51$ & $4.30$ & $3.46$ \\
        {\tt + rm$\_$inv} & $4.10$ & $5.14$ & $4.18$ & $3.46$ \\
        {\tt + enc$\_$norm} & $\textbf{4.08}$ & $\textbf{5.12}$ & $\textbf{4.16}$ & $\textbf{3.45}$ \\
        \hline
    \end{tabular}
    }
    \caption{Impact of cleaning on corpus perplexity.}
    \label{tab:norm_perplexity}
    \vspace{-3mm}
\end{table}

Although the normalization module has a clear advantage in preventing certain cases of intentional adversarial attacks \cite{kaing2024robust} during inference, it is still valuable to assess its impact on the corpus during model training. Intuitively, the normalization module reduces the encoding ambiguities and makes the corpus more predictable. A direct way to evaluate this effect is by computing perplexity on the corpus. This involves training a $5$-gram character-level frequency-based language model\footnote{\url{https://github.com/kpu/kenlm}} on a training corpus and calculating the perplexity on a held-out corpus.
We evaluated the normalization module on pre-trained corpora, including CC and Wiki, as well as on corpora from downstream tasks such as ALT and Lr-sum. For ALT and Lr-sum, we used their standard splits, with the dev set as the held-out corpus. For CC, we used the mC4 dev set as the held-out corpus. We split the Wiki corpus by using the Wikipedia corpus for training and Wikibook as the held-out set.
Table \ref{tab:norm_perplexity} compares the perplexity of the corpus when cleaned by {\tt rm\_inv} and then followed by {\tt enc\_norm}. As a result, the normalized corpora are indeed more predictable, particularly the corpora used for pre-training. Notably, invisible characters significantly hinder the predictability of the pre-trained corpora.

Additionally, we observed a reduction of around $20$k and $2$k unique vocabulary items\footnote{We segmented each corpus into words and extracted the unique vocabulary from each one.} for CC and Wiki, respectively, after applying {\tt enc\_norm}. This suggests that while ambiguous encodings are present, they are relatively rare in the corpora, though they could potentially cause issues during inference \cite{kaing2024robust}. Moreover, the differences in perplexity are relatively small for the fine-tuned corpora, which is expected, given that they originate from a single source and exhibit consistent text patterns.

\subsection{Quality of Tokenizers}\label{sec:tok_quality}

\begin{table}[t!]
    \centering
    \begin{tabular}{l|rr}
       Tokenizer & Fertility $\downarrow$ & Length Ratio $\downarrow$ \\
        \hline
        mBART50 & $9.27$ & $0.313$ \\
        Unigram & $\textbf{7.09}$ & $\textbf{0.238}$ \\
        PrahokBART & $7.81$ & $0.289$ \\
        \hline
    \end{tabular}
    \caption{Fertility of chunks and the average length ratio on ALT development set.}
    \label{tab:tokenizer_quality}
\end{table}

\begin{table*}[]
    \centering
    \begin{tabular*}{1.0\textwidth}{c}
        \parbox[c]{0cm}{\includegraphics[trim=0 5 20 0, width=15cm]{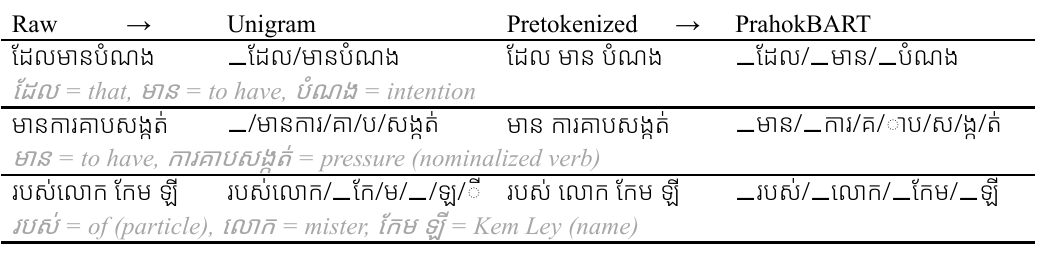}} \\
    \end{tabular*}
    \caption{Tokenization samples. Functional spaces in the third sample between the title, first and last name were excluded due to space limitation. mBART50 tokenized texts under Raw and PrahokBART tokenized those under word segmentation. `/' represents a delimiter space.}
    \label{tab:tokenization_samples}
\end{table*}

\begin{table}[t!]
    \centering
    \begin{tabular}{l|rr}
         Task & Unigram & PrahokBART \\
        \hline
        en$\rightarrow$km & $53.01$ & $\textbf{53.54}$\rlap{$^*$} \\
        km$\rightarrow$en & $52.93$ & $\textbf{53.26}$ \\
        TextSum & $18.07$ & $\textbf{26.23}$\rlap{$^\dag$} \\
        HeadGen & $20.49$ & $\textbf{22.92}$\rlap{$^\dag$} \\
        \hline
    \end{tabular}
    \caption{PrahokBART$_{big}$ performance with and without word segmentation, in terms of ChrF (first two rows) and Rouge-L (last two rows). $^\dag$ and $^*$ denote statistical significance with $p<0.01$ and $p<0.05$, respectively.}
    \label{tab:pretoken}
\end{table}

\begin{table}[t!]
    \centering
    \resizebox{\columnwidth}{!}{%
    \begin{tabular}{l|rrr}
         Task & mBART50 & Unigram & PrahokBART \\
        \hline
        en$\rightarrow$km & $0.78$ & $1.86$ & $\textbf{1.99}$ \\
        TextSum & $2.05$ & $1.51$ & $\textbf{2.63}$ \\
        HeadGen & $0.91$ & $0.86$ & $\textbf{1.00}$ \\
        \hline
    \end{tabular}
    }
    \caption{Quality of generated functional spaces in terms of BLEU score differences $\Delta S$.}
    \label{tab:real_space_gen}
\end{table}

\begin{table}[t!]
    \centering
    \begin{tabular}{l|rrr}
         Task & Low & Medium & High \\
         & ($39$M) & ($401$M) & ($4.2$B) \\
        \hline
        en$\rightarrow$km & $49.03$ & $52.48$ & $\textbf{53.54}$\rlap{$^\dag$} \\
        km$\rightarrow$en & $44.21$ & $49.69$ & $\textbf{53.26}$\rlap{$^\dag$} \\
        TextSum & $8.81$ & $22.81$ & $\textbf{26.23}$\rlap{$^*$} \\
        HeadGen & $12.66$ & $17.36$ & $\textbf{22.92}$\rlap{$^\dag$} \\
        \hline
    \end{tabular}
    \caption{PrahokBART$_{big}$ pre-trained on different data size, in terms of ChrF (first two rows) and Rouge-L (last two rows) scores. M and B denote million and billion of tokens, respectively. \revised{$^\dag$ and $^*$ denote statistical significance between Medium and High with $p<0.01$ and $p<0.05$, respectively.}}
    \label{tab:pretrain_data_size}
\end{table}

We assess the quality of our tokenizer using intrinsic metrics that measure how many subwords a tokenizer generates from a given text, such as fertility \cite{acs2019exploring, workshop2022bloom} and average length ratio \cite{zhang2022robust}. We compared our tokenizer against mBART50's tokenizer and a variant of our tokenizer that does not use word segmentation, denoted Unigram. Since the Unigram tokenizer and mBART50's tokenizer were trained on unsegmented texts, assessing their quality at the word level, which requires prior word segmentation, would be an unfair comparison with our tokenizer. Therefore, we measured fertility on phrase-like chunks. However, the average length ratio, which is the ratio between subwords and characters, is word-independent. 
Table \ref{tab:tokenizer_quality} shows that our tokenizer generated shorter sequences compared to mBART50's tokenizer, which is expected because mBART50 needs to cover other writing systems, such as Chinese, Japanese, Thai, and Myanmar, while ours covers only English and Khmer. 

Interestingly, word segmentation generated longer sequences when comparing our tokenizer with Unigram. To investigate further, we analyzed some samples tokenized by the Unigram and PrahokBART tokenizers, as shown in Table \ref{tab:tokenization_samples}. The Unigram tokenizer tends to generate larger chunks of tokens, resulting in shorter sequences, consistent with the results in Table \ref{tab:tokenizer_quality}. However, the Unigram tokenizer is not aware of the semantic units of words and often merges two separate words simply because they frequently occur together. Treating them as separate entities, however, is more beneficial for model learning.
In the first example, the Unigram tokenizer merged `to have' and `intention' into a single token, which seems linguistically reasonable because `to have intention' could be considered a compound phrase. However, this hinders the representation of `to have' when it appears with other words. In the second and third examples, the Unigram tokenizer merged `to have' with the nominal particle of `pressure' and also combined `of' with the prefix `mister,' both of which we believe negatively impact model performance. Incorporating word segmentation prevents such cases and produces linguistically motivated tokens. 

To further demonstrate the effectiveness of a linguistically motivated tokenizer, we conducted an extrinsic evaluation comparing our tokenizer with Unigram, as shown in Table \ref{tab:pretoken}. The results show that word segmentation, which produces linguistically motivated tokens, yields better performance, demonstrating the impact of utilizing a word segmentation module in the pre-trained models.

\subsection{Quality of Functional Space Generation}\label{sec:funcspaces}

We also quantify the model's performance in functional space generation by measuring the difference between score values with and without functional spaces, expressed as $\Delta S = S_{all} - S_{content}$. $S_{content}$, presented in the previous results, is the score measured without considering functional spaces, while $S_{all}$ includes them. A higher $\Delta S$ indicates better functional space generation. Intuitively, a higher $\Delta S$ signifies that more functional spaces match the reference.

Table \ref{tab:real_space_gen} shows that functional spaces generated by PrahokBART matched the reference the most. Compared to Unigram, we observe that word segmentation also contributed to better functional space generation. This could be because the functional spaces were treated as individual tokens rather than occasionally as prefixes. This finding aligns with the work of \citet{gow2022improving}, who identified that combining spaces with other tokens is problematic. They improved several NLU tasks involving complex words by treating spaces as individual tokens. We further found that this solution is effective for Khmer, where texts lack word boundaries and spaces serve as functional tokens, resulting in better functional space generation.

\subsection{Impact of Pre-training Data Size}

We believe that our models were trained on only a limited subset of the available data for Khmer. Additional data sources, such as various snapshots from Common Crawl or extensive internal datasets, could enhance the training process. As illustrated in Table \ref{tab:pretrain_data_size}, increasing the amount of training data improves downstream performance. We believe performance will be further boosted with more data, but this will require a larger model, according to the scaling law \cite{hoffmann2022training}. \revised{For instance, if the pre-training data are doubled to eight billion tokens, the model size would need to increase to approximately $400$ million parameters.}

\section{Conclusion}
We have introduced the first language-specific PS2S model tailored specifically for the Khmer language---PrahokBART. This model incorporates two linguistic modules during the preprocessing steps: normalization and word segmentation. PrahokBART demonstrates superior performance compared to its multilingual counterpart, mBART50, across three NLG tasks: MT, TextSum, and HeadGen. Our findings highlight the significant impact of linguistic modules in Khmer PS2S models. Specifically, normalization enhances the predictability of Khmer texts, while word segmentation generates linguistically motivated units, both of which improve downstream performance. Additionally, PrahokBART is capable of generating functional spaces that make output texts more natural. While PrahokBART shows promising results, there is still room for improvement. One key area for further development is the expansion of pre-training data. Although our model has achieved strong performance, we believe that with a larger pre-training dataset, PrahokBART’s potential could be further unlocked, leading to even better results in future iterations.

\section{Limitations}


Our model was trained exclusively on Khmer and English datasets, and the vocabulary is limited to these two languages. In other words, the model is beneficial only for downstream tasks that focus on the Khmer language or translation between English and Khmer. For other tasks, such as translation between Thai and Khmer, advanced techniques like vocabulary adaptation \cite{csaki2023efficiently} may be required to effectively utilize our models. Nevertheless, we believe this study provides valuable lessons for designing a language-specific PS2S model for languages that share similar characteristics with Khmer.

\section*{Acknowledgments}
\revised{We thank Jonas Belouadi for helping us plot Figure \ref{fig:plot_results}, and Haryo Akbarianto Wibowo and Ahmed Elshabrawy for their invaluable comments.}

\bibliography{custom}

\newpage
\appendix

\section{Implementation of Normalizer}\label{implement_norm}

For {\tt rm\_inv}, we detect and remove $29$ invisible characters listed in Table \ref{tab:inv_chars}. For {\tt enc\_norm}, we use normalization script provided by \cite{hosken2022khmer} written from page $40$ to page $42$.

\begin{table}[t]
    \centering
    \begin{tabular}{l|l|l|l}
        {\tt U+2063} & {\tt U+202A} & {\tt U+E007F} & {\tt U+200C} \\
        {\tt U+FEFF} & {\tt U+202C} & {\tt U+200F} & {\tt U+FE0E} \\
        {\tt U+E0067} & {\tt U+FE0F} & {\tt U+AD} & {\tt U+202D} \\
        {\tt U+180C} & {\tt U+E0065} & {\tt U+200E} & {\tt U+E01D3} \\
        {\tt U+17B5} & {\tt U+180B} & {\tt U+206E} & {\tt U+200B} \\
        {\tt U+180D} & {\tt U+E0062} & {\tt U+202B} & \\
        {\tt U+E006E} & {\tt U+2060} & {\tt U+17B4} & \\
        {\tt U+200D} & {\tt U+180E} & {\tt U+2061} & \\
    \end{tabular}
    \caption{List of invisible characters to remove.}
    \label{tab:inv_chars}
\end{table}

\section{Units of Text}
Khmer texts are written without spaces between words, and spaces, which we call functional spaces are used as commas or simply for readability. These spaces are commonly inserted between clauses or phrases and sometimes between words such as conjuncts and English words.
In this paper, we simply refer to such units as `phrases' because the units are larger than words in general.
Furthermore, `words' refers to those units segmented by a word segmenter, and `subwords' refers to those tokenized by a subword tokenizer.
There are two scenarios of subword tokenization in this paper: performing subword tokenization on phrases or on words. 
In Sections \ref{sec:tok_quality} and \ref{sec:funcspaces}, `Unigram' refers to the first scenario, where subword tokenization is performed on phrases, and `PrahokBART', our model in which word segmentation is performed before subword tokenization, refers to the second scenario.

\section{Quality Analysis}
\begin{table*}[t!]
    \centering
    \begin{tabular*}{1.0\textwidth}{l|l}
        \hline
        \parbox[t]{2mm}{\multirow{1}{*}{\rotatebox[origin=c]{90}{km$\rightarrow$en}}} & \parbox[c]{0cm}{\includegraphics[trim=5 5 0 0, width=15.5cm]{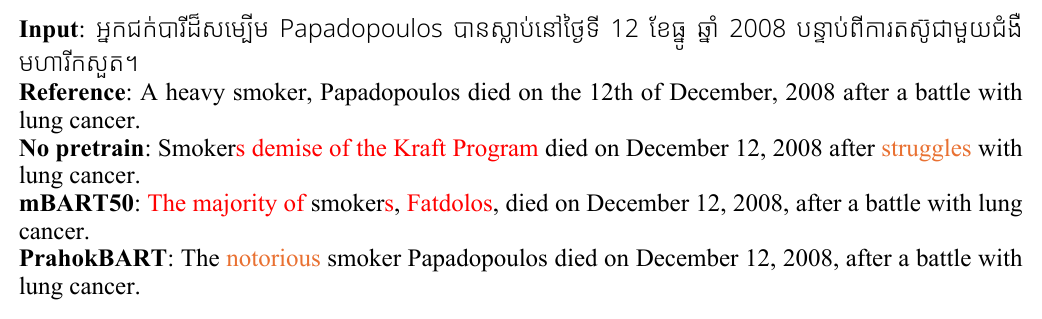}} \\
        \hline
        \parbox[t]{2mm}{\multirow{1}{*}{\rotatebox[origin=c]{90}{en$\rightarrow$km}}} & \parbox[c]{0cm}{\includegraphics[trim=5 5 0 0, width=15.5cm]{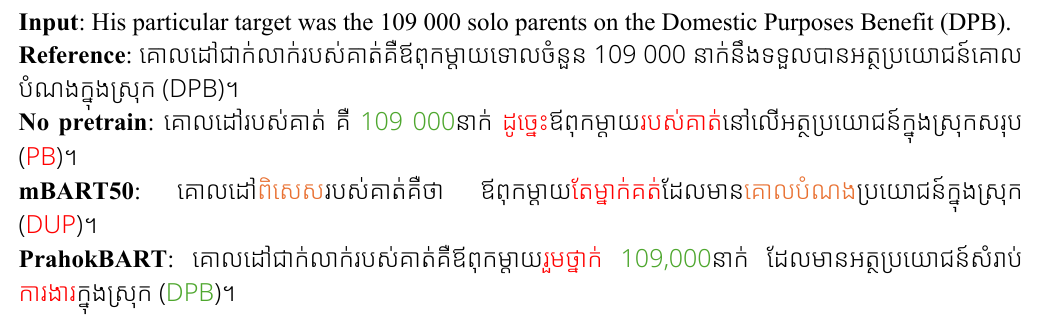}} \\
        \hline
        \parbox[t]{2mm}{\multirow{1}{*}{\rotatebox[origin=c]{90}{TextSum}}} & \parbox[c]{0cm}{\includegraphics[trim=5 5 0 0, width=15.5cm]{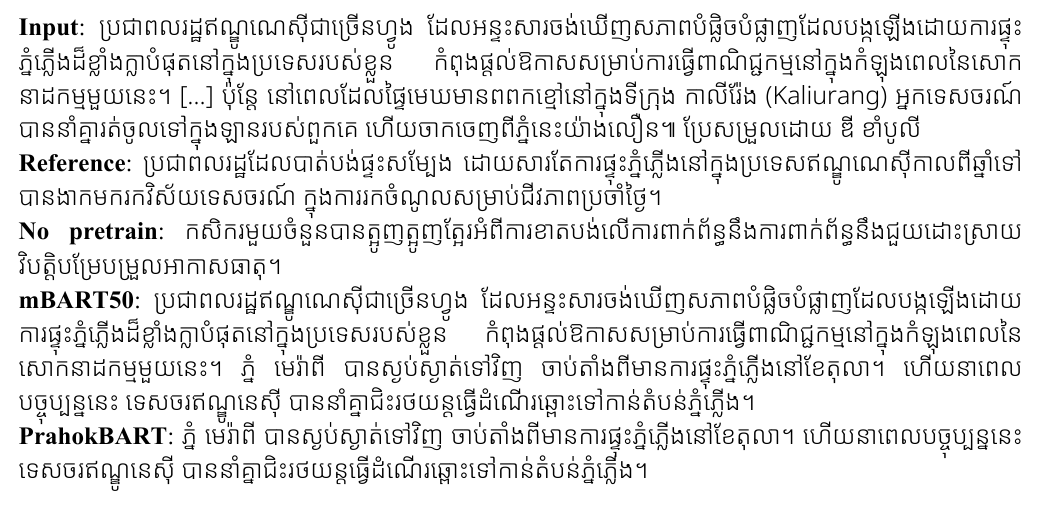}}  \\
        \hline
        \parbox[t]{2mm}{\multirow{1}{*}{\rotatebox[origin=c]{90}{HeadGen}}} & \parbox[c]{0cm}{\includegraphics[trim=5 0 0 0, width=15.5cm]{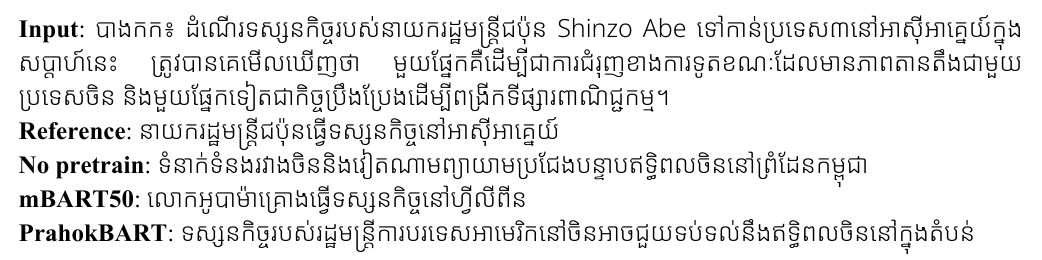}}  \\
        \hline
    \end{tabular*}
    \caption{Additional samples for en$\rightarrow$km, TextSum, and HeadGen. Texts in \textcolor{red}{red} are incorrect translations and those in \textcolor{orange}{orange} are acceptable.}
    \label{tab:additional_samples}
\end{table*}

We randomly sampled an example for each task from a pool where the outputs generated by the Random baseline had low scores, and the references were short. By doing this, we aimed to assess the improvement brought by pre-trained models on those samples, as shown in Table \ref{tab:additional_samples}.

For the MT task, we observe that PrahokBART excels at copying key words from the input to the target translation, especially proper nouns in English, compared to the baselines. This includes abbreviations in English and even numbers. In contrast, mBART50 seems to struggle to translate or copy these keywords effectively.

For TextSum, the task appears to be quite challenging, and none of the models generated an output that semantically matches the reference, apart from a few matching words. As seen in the sample, mBART50 simply copied the first two sentences, while our model only included the second sentence. Similar to TextSum, HeadGen is also challenging, as the models struggle to generate a headline that accurately describes the intent of the article. Although some words matched, the main keywords were often incorrectly generated by all the models. This analysis highlights the need for further research on TextSum and HeadGen for Khmer.

\end{document}